\documentclass[letterpaper, 10 pt, conference]{ieeeconf}  
\usepackage{times}

\IEEEoverridecommandlockouts                              

\usepackage{multicol}
\usepackage[bookmarks=true]{hyperref}
\usepackage{graphicx}
\usepackage{amsfonts}
\usepackage{amsmath,amssymb} 
\usepackage{algorithm}
\usepackage[noend]{algpseudocode}
\usepackage{multirow}
\usepackage{subfig}
\usepackage{caption}
\usepackage{soul} 
\usepackage{bchart}
\usepackage{newfloat}
\usepackage{bbm}
\usepackage{pgfplots}
\usepackage{fancyhdr}
\pgfplotsset{width=9cm,compat=1.8}

\usepackage{ifthen}
\usetikzlibrary{matrix,calc}

\DeclareFloatingEnvironment[
   fileext=loc
]{chart}

\usepackage[colorinlistoftodos,prependcaption,textsize=tiny]{todonotes}
\usepackage[normalem]{ulem}
\begin{document}

\title{Synthetic-to-Real Domain Adaptation for Action Recognition: \\A Dataset and Baseline Performances}

\author{Arun V. Reddy$^{1,2*}$, Ketul Shah$^{1*}$, William Paul$^{2}$, Rohita Mocharla$^{2}$, Judy Hoffman$^{3}$ \\Kapil D. Katyal$^{2}$, Dinesh Manocha$^{4}$, Celso M. de Melo$^{5}$, Rama Chellappa$^{1}$
\thanks{$^{1}$Johns Hopkins University, Dept. of Electrical \& Computer Engineering, Baltimore, MD, USA.}%
\thanks{$^{2}$Johns Hopkins University Applied Physics Lab, Laurel, MD, USA.}%
\thanks{$^{3}$Georgia Institute of Technology, Atlanta, GA, USA.}%
\thanks{$^{4}$University of Maryland, College Park, MD, USA.}%
\thanks{$^{5}$Army Research Lab, Adelphi, MD, USA.}%
\thanks{$^{*}$These authors contributed equally: {\tt\small areddy24@jhu.edu, kshah33@jhu.edu}}
}
\maketitle

\begin{abstract}

Human action recognition is a challenging problem, particularly when there is high variability in factors such as subject appearance, backgrounds and viewpoint. While deep neural networks (DNNs) have been shown to perform well on action recognition tasks, they typically require large amounts of high-quality labeled data to achieve robust performance across a variety of conditions. Synthetic data has shown promise as a way to avoid the substantial costs and potential ethical concerns associated with collecting and labeling enormous amounts of data in the real-world. However, synthetic data may differ from real data in important ways. This phenomenon, known as \textit{domain shift}, can limit the utility of synthetic data in robotics applications. To mitigate the effects of domain shift, substantial effort is being dedicated to the development of domain adaptation (DA) techniques. Yet, much remains to be understood about how best to develop these techniques. In this paper, we introduce a new dataset called Robot Control Gestures (RoCoG-v2). The dataset is composed of both real and synthetic videos from seven gesture classes, and is intended to support the study of synthetic-to-real domain shift for video-based action recognition. Our work expands upon existing datasets by focusing the action classes on gestures for human-robot teaming, as well as by enabling investigation of domain shift in both ground and aerial views. We present baseline results using state-of-the-art action recognition and domain adaptation algorithms and offer initial insight on tackling the synthetic-to-real and ground-to-air domain shifts. Instructions on accessing the dataset can be found at \url{https://github.com/reddyav1/RoCoG-v2}.
\end{abstract}


\fancypagestyle{firstpage}{
  \fancyhf{} 
  \renewcommand{\headrulewidth}{0pt} 
  \lfoot{} 
  \cfoot{} 
  \rfoot{\textit{Distribution Statement A: Approved for public release. Distribution is unlimited.}} 
}

\thispagestyle{firstpage}

\begin{figure}[]
	\centering
    \includegraphics[width=.95\columnwidth]{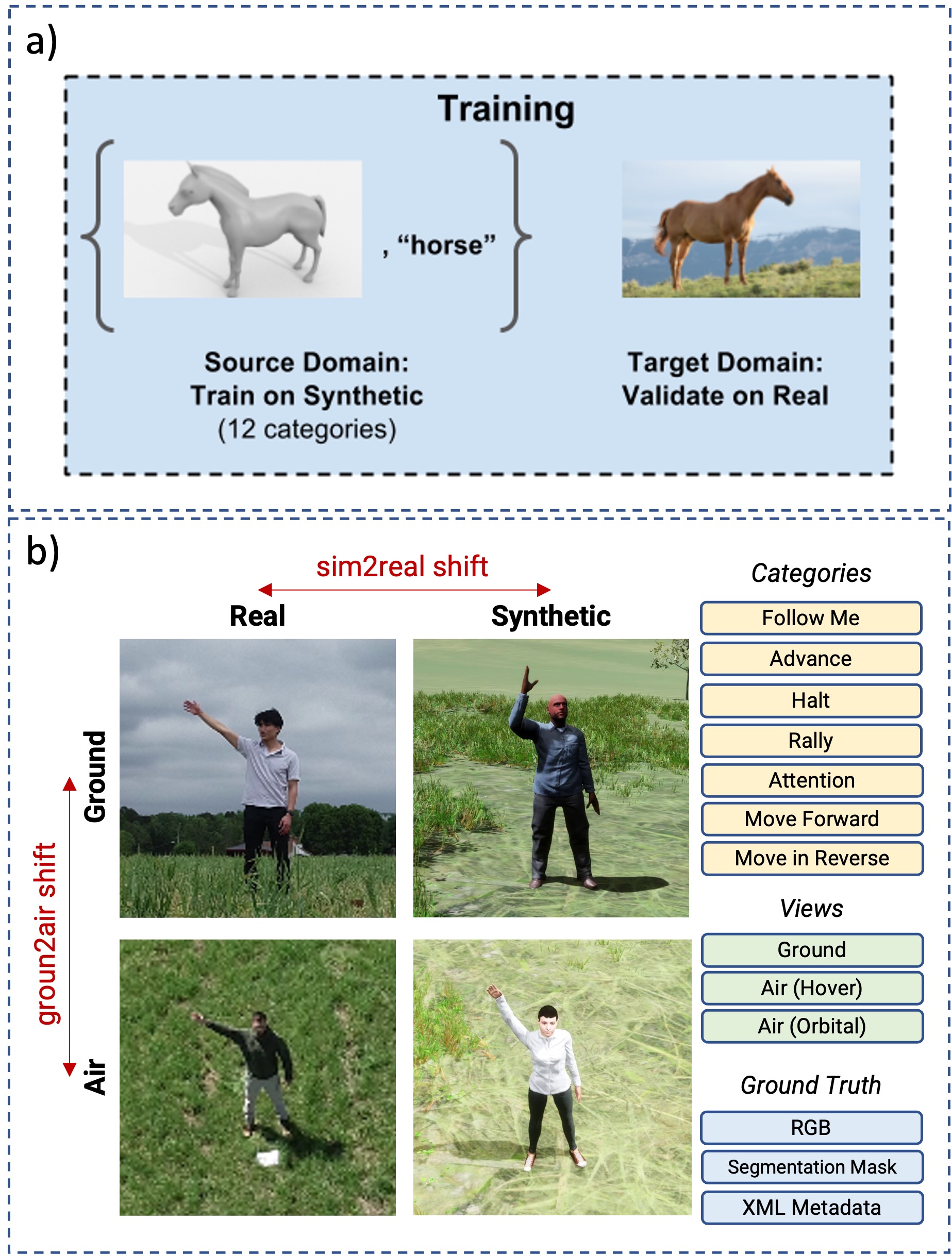}
    \caption{a): Many existing domain adaptation datasets, including VisDA~\cite{visda2017}, focus on image classification or semantic segmentation. b): In this paper, we investigate domain adaptation techniques across two domains: synthetic-to-real and ground-to-air, focused on human action recognition.}
	\label{fig:overview}
 \vspace{-0.2cm}
\end{figure}

\section{Introduction}

Human action recognition from ground-based cameras and/or airborne videos (e.g., from unmanned aerial vehicles) is a challenging problem and has received much attention in the computer vision and robotics literature. Action recognition can enhance human-agent teaming through gesture communication, help search and rescue efforts, enable learning by social imitation, and increase social awareness (e.g., autonomous driving). In recent years, remarkable performance using deep neural networks (DNNs) has been obtained for action classification~\cite{simonyan2014two, karpathy2014large, negin2016human, yue2015beyond, carreira2017quo, hou2017tube, saha2017amtnet,choutas2018potion, weinzaepfel2021mimetics, yan2019pa3d}. While classification accuracy is critical, equally important are data efficiency and robustness to varying viewpoints.

Traditionally, action recognition from videos collected by ground-based and airborne cameras has been treated as two separate problems. There are many instances where the training dataset is collected from ground-based cameras and the test data is acquired by unmanned aerial vehicles (UAVs), or vice versa. This problem is challenging due to the large appearance and geometry-based domain shift between the two acquisition conditions. Over the last decade, new classes of methods known as domain adaptation have been proposed to address such shifts. Domain adaptation (DA) seeks to close the gap that occurs when DNNs are applied to a target distribution (e.g., real-world video) that has different characteristics than the source training distribution (e.g., synthetic video). To transfer knowledge across domain-shifted data distributions, it would be beneficial to have a feature representation that effectively reduces biases contributing to the domain shift. A variant of domain adaptation is unsupervised domain adaptation (UDA), which attempts to adapt models trained on labeled source domain data using unlabeled data from the target domain. Given the high cost of labeling data, the prospect of relying mostly on simulation and incorporating a limited amount of real data, without the need for labeling, is attractive in many applications. Over the last decade, domain adaptation methods have been designed for many computer vision problems (such as object recognition, semantic segmentation, action recognition, etc.) using datasets such as Office Dataset~\cite{saenko2010adapting} and DomainNet~\cite{peng2019moment}. However, in spite of this progress, much still remains to be understood about how to develop general solutions to close the source-to-target gaps. 

One domain shift of particular relevance is that between synthetic and real data. Synthetic data is becoming increasingly critical to sustaining the deep learning revolution~\cite{sankaranarayanan2018learning, hoffman2018cycada, deMeloEtAl22, Nikolenko19}. Recent years have experienced remarkable performance of DNNs in many areas of robotics which include visual perception, question and answer tasks, navigation, and control. However, training a DNN typically requires large amounts of high-quality labeled data, which is often the main bottleneck in the model development process. Synthetic data offers a potential solution to this challenge since it is usually easier to acquire, controllable, pre-annotated, less expensive, inexhaustible, and can avoid practical and ethical issues (e.g., security and privacy concerns). However, a central challenge occurs when attempting to transfer a DNN trained on synthetic data to a real domain.

As seen in Fig. \ref{fig:overview}, in this paper we consider the problem of adapting action recognition models trained on synthetic data for use on real data. In addition to the synthetic-to-real shift, we also consider the domain shift due to differences in viewpoint.

In summary, the paper makes the following contributions:
\begin{itemize}
  \item A novel dataset (RoCoG-v2) for the study of domain adaptation solutions, composed of synthetic and real videos from seven gesture classes, from both the ground and air perspectives;
  \item Baseline experiments on the dataset using state-of-the-art gesture recognition and domain adaptation algorithms;
  \item Insight on the synthetic-to-real and ground-to-air domain adaptation challenges, which have considerable practical relevance for robotics applications.
\end{itemize}

\section{Related Work}
\paragraph{Synthetic Datasets} 

The use of synthetic data to augment difficult-to-acquire real data is an active area of research with many promising results \cite{deMeloEtAl22,Nikolenko19}. The impact of using synthetic data is particularly relevant in robotics where real-world data can be challenging to collect, expensive to label and is often not robust to changing environments and contexts. For this reason, many synthetic datasets have been introduced in recent years including VisDA~\cite{visda2017} for domain adaptation techniques in image classification and segmentation, synthetic datasets for ground robots~\cite{9649285,10.1016/j.robot.2019.103336,7487210}, aerial robots~\cite{DBLP:journals/corr/abs-2112-12252} and action recognition including \cite{DBLP:journals/corr/abs-2007-11118,9206624,9340728}. Action recognition from aerial videos has also been studied extensively~\cite{nguyen2022state} and aided by datasets like~\cite{li2021uav, perera2018uav, barekatain2017okutama}.~\cite{choi2020unsupervised} introduces the NEC-DRONE dataset for ground-to-air domain adaptation. Further, many simulators have also been developed to support generation of synthetic data. These simulators, including CARLA~\cite{Dosovitskiy17}, GTA~\cite{Richter2016PlayingFD}, NVIDIA ISAAC/Omniverse~\cite{2021_28dd2c79}, Habitat~\cite{szot2021habitat,habitat19iccv}, AI2Thor~\cite{RoomR}, and iGibson~\cite{li2021igibson} allow users to create virtual environments that provide error-free, ground truth annotations.

\paragraph{Synthetic-to-Real Transfer} 
Many studies have also explored synthetic-to-real transfer by leveraging synthetic data generated from simulation environments to pretrain a neural network, thereby reducing the amount of data needed in real-world settings~\cite{10.1007/978-3-031-04870-8_16,DBLP:journals/corr/abs-2012-03806}. Deep Adaptive Networks (DAN)~\cite{10.5555/3045118.3045130} uses a Hilbert space representation for the embedding representations to match different domain distributions. Adversarial Discriminative Domain Adaptation (ADDA)~\cite{10.1145/3357384.3357918} and  Generate to Adapt~\cite{8578985} are domain adaptation techniques that use unsupervised data to learn a joint feature space between the source and target distributions. CyCADA~\cite{hoffman2018cycada} extends existing adversarial domain adaptation techniques by adding a cycle-consistency constraint for unsupervised adaptation from synthetic to real-world driving domains. Randomized-to-Canonical Adaptation Networks (RCANs)~\cite{inproceedingsgrasp}, GraspGAN~\cite{Bousmalis2018UsingSA}, RetinaGAN~\cite{9561157}, Closing the Sim-to-Real Loop~\cite{8793789}, and the work described in \cite{8793561} leverage simulation and domain adaptation techniques for robot manipulation and grasping. VR-Goggles for Robots~\cite{DBLP:journals/ral/ZhangTYX0BB19} focuses on converting real images back to simulation for visual control of robots in indoor and outdoor environments. The authors of \cite{sim2real_outdoor} use synthetic-to-real techniques to improve mobile robot control policies on uneven, complex outdoor environments. 
Two UDA methods of particular relevance in this paper are Domain Adversarial Neural Network (DANN)~\cite{DANN} and Contrastive Conditional domain Alignment (CO\textsuperscript{2}A)~\cite{CO2A}, which are used in baseline experiments on the RoCoG-v2 dataset (see Sec.~\ref{sec:baseline_experiments}). DANN employs an adversarial approach to domain alignment with the use of a gradient reversal layer. CO\textsuperscript{2}A, on the other hand, uses contrastive losses at both the frame- and video-level to perform domain adaptation.

RoCoG-v2 is unique among its counterparts in that it consists of two different types of domain shifts (synthetic-to-real and ground-to-air). It also offers a greater volume of videos (107,478 in total) than comparable synthetic-to-real datasets like Mixamo-Kinetics (which contains 36,195 videos) \cite{CO2A}. In addition, RoCoG-v2 includes videos rendered using MoCap sequences, and thus exhibits greater motion realism than RoCoG-v1 \cite{deMeloEtAl20}. By releasing this dataset, our goal is to further explore the benefits of synthetic data in improving gesture recognition for robot control.

\section{Dataset}

\begin{figure*}
  \centerline{\includegraphics[width=0.95\textwidth]{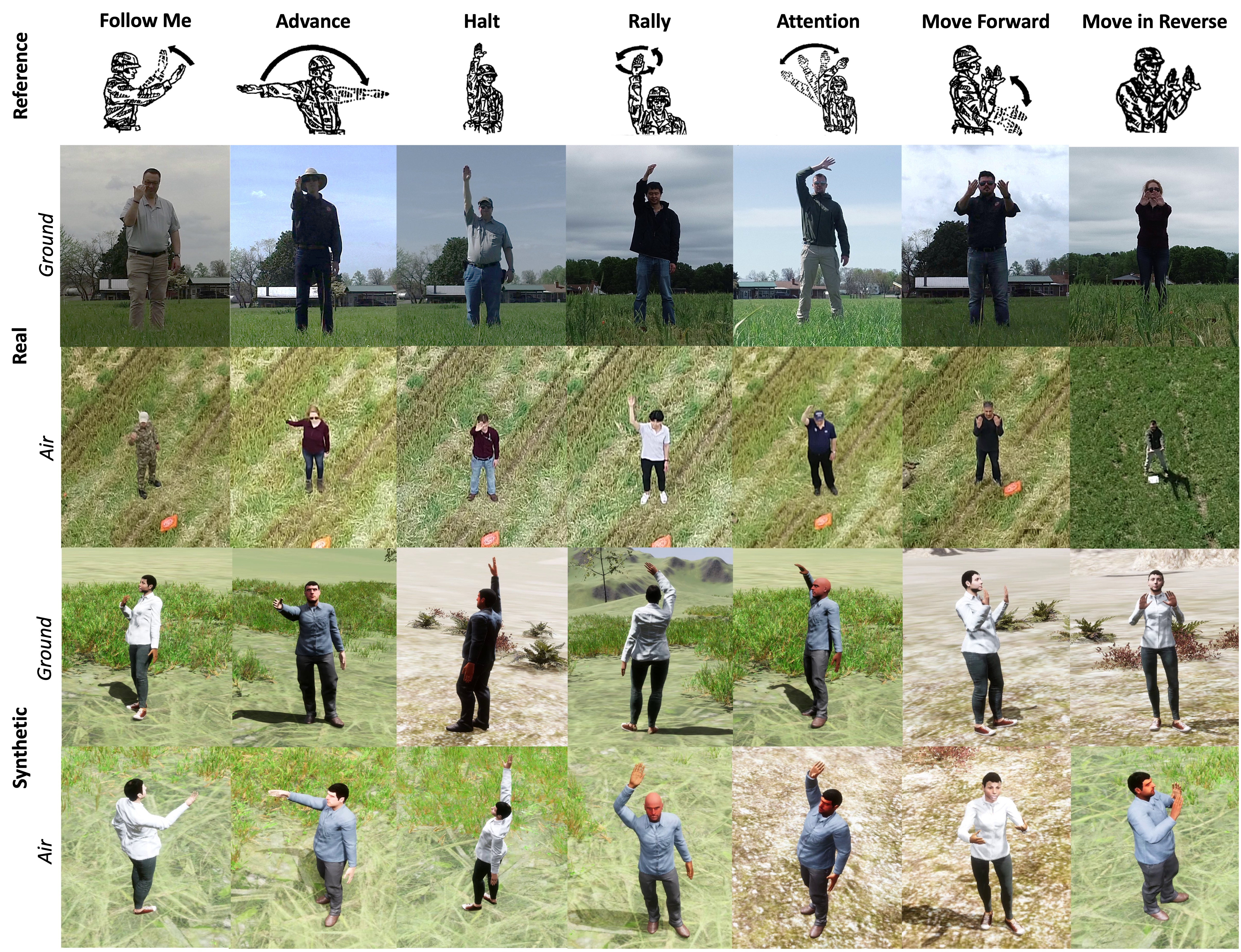}}
\caption{The dataset consists of real and synthetic videos across the seven gesture classes, from both ground and air perspectives.}
\label{fig:dataset}       
\end{figure*}

While RoCoG-v2 is useful for the general study of domain adaptation, it focuses in particular on two challenges of high practical relevance: the synthetic-to-real and ground-to-air domain shifts in gesture recognition. The dataset consists of real and synthetic videos from seven action classes, viewed from both the ground and air perspectives (Fig. \ref{fig:dataset}). The actions consist of seven control gestures retrieved from the U.S. Army Field Manual \cite{USArmy87} (Fig. \ref{fig:dataset}, 1\textsuperscript{st} row): \texttt{follow me}, \texttt{advance}, \texttt{halt}, \texttt{rally}, \texttt{attention}, \texttt{move forward}, and \texttt{move in reverse}. It is worth noting that some of the gestures have similar appearance (e.g., ~\texttt{move forward} and \texttt{move in reverse}), which presents an additional challenge to action recognition algorithms. This dataset can be seen as a more general, considerably improved version of RoCoG-v1 \cite{deMeloEtAl20}. 

The real videos were collected in two outdoor locations with 10 adult subjects (9 males and 1 female, with diverse age ranges and clothing types). The recordings were performed with DJI M210 drones equipped with Zenmuse X4S cameras. For the ground perspective, the drone was placed on the ground at a 10-meter distance and oriented towards the subject. For the air perspective, the drone was placed 15 meters away from the subject at a zenith angle of 45$^{\circ}$. We recorded subjects performing the gestures both with the drone hovering in a static location and while orbiting the subject. The latter configuration supports studying the algorithm's robustness to camera ego motion. However, for simplicity, the experiments presented in the paper focus exclusively on the comparison between the ground and hovering configurations. Subjects were given brief instructions about the gestures and were asked to perform several repetitions of each gesture ``in whatever manner seemed natural to them." The instructions lacked detail by design to capture diversity in the performance of the gestures (e.g., the number of repetitions in the \texttt{rally} or \texttt{follow me} gestures). 

The synthetic data was generated using a custom simulator built on Unity, a commercial game engine. We systematically varied parameters for the scene and, for each scene, the parameters for gesture performance (Table \ref{tab:synth_data_parameters}). To increase the visual realism of the data, we used high-quality 3D assets from online repositories and rendered the scenes using a high definition rendering pipeline. The gestures were animated using two types of animation techniques with different qualities of motion realism: \textit{skeleton-based animation}, where the gesture is animated manually by an artist, and \textit{motion capture} (\textit{mocap}), where motion data is recorded, using markers, from gesture performances by a separate sample of human subjects. Each frame of the video was labeled with the corresponding gesture class for periods when a gesture was being performed. Though not leveraged in the experiments reported in this paper, the dataset also contains semantic segmentation masks for every frame. In total, nearly 107K synthetic videos were rendered across the two views. 

The details of the full dataset by video type and split are shown in Table \ref{tab:dataset_stats}. To better assess generalization, we used scene-based partitioning to produce training, validation, and testing partitions for each domain. For each synthetic data domain, 40 scenes were randomly selected for training, and 8 were used for validation. For real data, 4 subjects were carefully chosen to form the test set. The set of 4 test subjects, which includes the only female subject, captures variation in background location, subject body build and subject skin color. The remainder of the subjects were used during training.

\begin{table}
\caption{Synthetic Data Parameters.}
\label{tab:synth_data_parameters}
\begin{center}
\begin{tabular}{p{1.6cm}p{6cm}}
\hline\noalign{\smallskip}
\textbf{Parameter} & \textbf{Range}  \\
\noalign{\smallskip}\hline\noalign{\smallskip}
\multicolumn{2}{c}{\textit{Scene Parameters}}\\
Terrain & Grassy, Desert \\
Character & Male, Female \\
Lighting & Configuration 1, Configuration 2 \\
Camera & 30$^{\circ}$, 90$^{\circ}$, 150$^{\circ}$, 210$^{\circ}$, 270$^{\circ}$, 330$^{\circ}$ \\
Perspective & Ground, Air \\
\noalign{\smallskip}\hline\noalign{\smallskip}
\multicolumn{2}{c}{\textit{Gesture Parameters}}\\
Class & Follow me, Advance, Halt, Rally, Attention, Move forward, Move in reverse \\
Type & Animation, Motion capture \\
Variation & Three variations per gesture \\
Speed & 1$\times$, 1.25$\times$ \\
Race & Caucasian, African-American, East Indian \\
Thickness & Thin, Thick \\
\noalign{\smallskip}\hline
\end{tabular}
\end{center}
\end{table}
\begin{table}[h]
\caption{Number of videos of each type in RoCoG-v2. }
  \centering
   \renewcommand{\arraystretch}{1.1}
  \begin{tabular}{|c|c|c|c|c|c|}
  \hline
   & & \textbf{Train} &  \textbf{Val} & \textbf{Test} &  \textbf{Total} \\

  \hline
  \multirow{2}{*}{\textbf{Synthetic}} &  \textbf{Ground} & 44,510 & 8,928 & - & 53,438 \\
  \cline{2-6} 
  & \textbf{Air} & 44,640 & 8,918 & - & 53,558 \\
  \hline
   \multirow{2}{*}{\textbf{Real}} &  \textbf{Ground} & 204 & - & 100 & 304 \\
  \cline{2-6}
  & \textbf{Air} & 87 & - & 91 & 178 \\
  \hline
 
  \end{tabular}

  \label{tab:dataset_stats}
\end{table}

\section{Baseline Experiments}
\label{sec:baseline_experiments}
\subsection{Algorithms}
\label{sec:baseline_algorithms}

Baseline results are provided using action recognition models trained on only the source domain (``Source Only") and target domain (``Target Only"), ideally representing the worst-case and best-case respectively for performance on the target domain. We also attempt to address the domain gap using two unsupervised domain adaptation algorithms--DANN \cite{DANN}, which is a classic adversarial approach to UDA, and CO\textsuperscript{2}A \cite{CO2A}, which represents the current state-of-the-art algorithm for synthetic-to-real video domain adaptation. 

DANN performs domain adaptation by encouraging extraction of domain-invariant features. This is achieved by maximizing the loss of a \textit{domain classifier} network (which attempts to distinguish between the source and target domains), while minimizing the loss of a \textit{label predictor} network (which performs the main classification task on source domain data). We implement DANN for videos using separate domain classifiers at the frame- and video-level.

CO\textsuperscript{2}A divides the video into smaller segments denoted as clips, runs the backbone on each clip, and aggregates the clip-level features using a form of attention into video-level features used for classification. Clip-level features are also projected and utilized in either a supervised contrastive loss or self-supervised contrastive loss for source and target data respectively. To perform domain adaptation, target projected clip features are pseudo-labeled by the network's current prediction, and an inter-domain contrastive loss is used where clip features from different domains, but with the same label, are considered as positive pairs to enforce alignment.

\begin{table*}
\centering
\caption{Baseline UDA results (top-1 accuracy \%) on the four domain shifts of interest in RoCoG-v2.}
\label{tab:results}
\begin{center}
\begin{tabular}{ c c c c c c } 
\hline\noalign{\smallskip}
\textbf{Method} & \textbf{Architecture} & \multicolumn{1}{c}{{$G_S \rightarrow G_R$}} & \multicolumn{1}{c}{$A_S \rightarrow A_R$} & \multicolumn{1}{c}{$G_R \rightarrow A_R$} & \multicolumn{1}{c}{$G_S \rightarrow A_R$} \\
\hline\noalign{\smallskip}
\multirow{2}{6em}{Source Only} & I3D & $80.3 \pm 5.5$ & $48.0 \pm 6.7$ & $52.4 \pm 9.3$ & $41.1 \pm 5.5$  \\ 
& X3D & $75.3 \pm 0.6$ & $34.5 \pm 6.6$ & $54.2 \pm 6.0$ & $34.1 \pm 4.0$ \\ 
\hline\noalign{\smallskip}
\multirow{2}{6em}{DANN} & I3D & $69.7 \pm 5.5$ & $45.4 \pm 6.4$ & $53.5 \pm 1.7$ & $41.4 \pm 4.4$ \\ 
& X3D & $79.0 \pm 3.6$ & $61.2 \pm 8.3$ & $49.8 \pm 1.7$ & $64.8 \pm 9.6$ \\ 
\hline\noalign{\smallskip}
\multirow{2}{6em}{CO\textsuperscript{2}A} & I3D & $70.3 \pm 0.6$ & $60.4 \pm 4.4$ & $56.4 \pm 3.5$ & $45.1 \pm 4.4$ \\ 
& X3D & $74.0 \pm 3.0$ & $63.0 \pm 2.8$ & $56.4 \pm 8.1$ & $32.4 \pm 0.8$\\ 
\hline\noalign{\smallskip}
\multirow{2}{6em}{Target Only} & I3D & $83.0 \pm 2.7$ & $68.1 \pm 6.6$ & $68.1 \pm 6.6$ & $68.1 \pm 6.6$ \\ 
& X3D & $87.0 \pm 2.7$ & $70.3 \pm 2.9$ & $70.3 \pm 2.9$ & $70.3 \pm 2.9$  \\ 
\hline\noalign{\smallskip}
\end{tabular}
\end{center}
\end{table*}
\subsection{Experimental Setup}
\label{sec:experimental_setup}
Given the large discrepancy in field-of-view between the real and synthetic videos, we pre-process the real videos to equalize the scale of the human within the frame. Specifically, we localize and crop around the human subject in the real videos using bounding boxes produced by the Detectron2 framework \cite{wu2019detectron2}.

For training and evaluating action recognition models, we use the mmaction2 \cite{2020mmaction2} framework for all experiments. We choose two backbones: the widely used I3D \cite{carreira2017quo} (ResNet50-based) model and the more recent X3D \cite{feichtenhofer2020x3d} (M) model. The input to the model is a 16-frame clip of resolution 256$\times$256, resizing the video frames if necessary, for all methods expect CO\textsuperscript{2}A, which takes four 16-frame clips as input. For data augmentation, we use random horizontal flip, scale augmentations and RandAug~\cite{NEURIPS2020_d85b63ef} for all methods except CO\textsuperscript{2}A, where we use random horizontal flip, scale augmentation and color jitter, to be consistent with the original implementation. For all experiments, we initialize our models with Kinetics-400 pre-trained weights.

\subsection{Results}
\label{sec:results}

The results are organized according to four domain shifts: $G_S \rightarrow G_R$, focused on the synthetic-to-real shift for ground data; $A_S \rightarrow A_R$, focused on the synthetic-to-real shift for air data; $G_R \rightarrow A_R$, focused on the ground-to-air shift for real data only; and, $G_S \rightarrow A_R$, which features both the synthetic-to-real and ground-to-air shifts. These represent the typical cases for many applications; however, as noted in Section \ref{sec:conclusion}, this dataset can also be used to study other domain shifts.

For each domain shift, we report top-1 accuracy when training on the source domain only, target domain only, and when applying the two domain adaptation algorithms. Given the significant variation in results, which likely stems from the small test set sizes, we conduct three trials of each experiment and report the mean and standard deviation across them. The top-1 accuracy results are reported in Table \ref{tab:results} and run-averaged confusion matrices are shown in Fig~\ref{fig:confusion}.

\section{Discussion}
\label{sec:discussion}

\subsection{Ground (Synthetic) $\rightarrow$ Ground (Real)}
\label{sec:ground_syn_to_ground_real}
Experiments in this setting indicate that our synthetic ground data alone is reasonably effective in training models that perform well on real ground data. When using an I3D backbone, the source only accuracy (80.3\%) is not far off from the target only accuracy (83.0\%). While this discrepancy is larger with X3D backbone (75.3\% source only vs. 87.0\% target only), the source only accuracy is still respectable. This observation suggests the utility of realistic synthetic data, possibly even without domain adaptation. 
Interestingly, the results when employing UDA algorithms do not show consistent improvement over the source only accuracy. A possible explanation for this is the relatively low severity of the domain gap in this setting compared to the other settings evaluated.

\subsection{Air (Synthetic) $\rightarrow$ Air (Real)}
\label{sec:air_syn_to_air_real}
Our findings suggest that action recognition from the air can be more challenging than from the ground, as revealed by the direct comparison of the source only ($A_S \rightarrow A_R$: 34\% vs. $G_S \rightarrow G_R$: 75\%, for X3D) and target only ($A_S \rightarrow A_R$: 70\% vs. $G_S \rightarrow G_R$: 87\%, for X3D) baselines. This is possibly due to the loss of (geometric) information as the angle of view approaches the nadir angle.

The results show, once again, evidence of synthetic-to-real domain shift ($\sim$36\% difference between source only and target only when using X3D). The magnitude of the synthetic-to-real performance gap is greater for the air viewpoint than for the ground viewpoint, which may also reflect the increased difficulty in perceiving action from the air. In this case, domain adaptation techniques generally improved the accuracy (e.g., CO\textsuperscript{2}A with the X3D backbone led to $\sim$29\% boost in performance, when compared to the source only baseline), while still leaving room for future improvement. 

\subsection{Ground (Real) $\rightarrow$ Air (Real)}
\label{sec:ground_real_to_air_real}
The results show, as expected, a drop in performance when shifting from the ground to the air perspective ($\sim$16.1\% drop with X3D). The domain adaptation techniques show only modest improvement in performance over the source only baseline, which may suggest that other techniques that explicitly address the viewpoint shift may be needed to close the gap.

\subsection{Ground (Synthetic) $\rightarrow$ Air (Real)}
\label{sec:ground_syn_to_air_real}
The results in this setting reveal the intuitive observation that the domain gap is largest ($\sim$36.2\% difference between source only and target only with X3D) when both synthetic-to-real and ground-to-air domain shifts are present. Domain adaptation methods show only minimal improvements for this setting, suggesting the difficulty of addressing multiple types of domain shift simultaneously. Curiously, DANN with X3D backbone leads to substantial improvement, albeit with a large variance.

\subsection{Class Confusion Analysis}
The confusion matrices in Figure \ref{fig:confusion} show how errors are distributed among the gesture classes. One observation is that certain classes are commonly confused in all scenarios (e.g., \texttt{Move Forward} and \texttt{Move in Reverse}). For the $G_S \rightarrow G_R$ adaptation scenario (Fig~\ref{fig:confusion}: Top Left), we see that \texttt{Advance} is misclassified most often as \texttt{Attention}, which is perhaps not surprising given that the two gestures share some characteristics when observed from a frontal ground view. Some classes (e.g., \texttt{Attention} and \texttt{Rally}) are confused more frequently from the aerial view than from the ground view, leading to confusion as seen in Fig~\ref{fig:confusion}: Top Right. This type of class-wise analysis may play a role in the development of more specialized domain adaptation techniques.

\def\myConfMat{{
{286,	47,	0,	24,	0,	0,	0},
{357,	953,	0,	166,	0,	67,	0},
{95,	0,	1000,	0,	167,	0,	0},
{24,	0,	0,	381, 0,	0,	0},
{143,	0,	0,	0,	833,	22,	0},
{24,	0,	0,	0,	0,	711,	0},
{71,	0,	0,	429,	0,	200,	1000},
}}

\def\classNames{{"Advance","Attention","Rally","Move Forward","Halt", "Follow Me", "Move in Reverse"}} 

\def\numClasses{7} 

\def\myScale{0.70} 

\def\scriptsize{0.25}

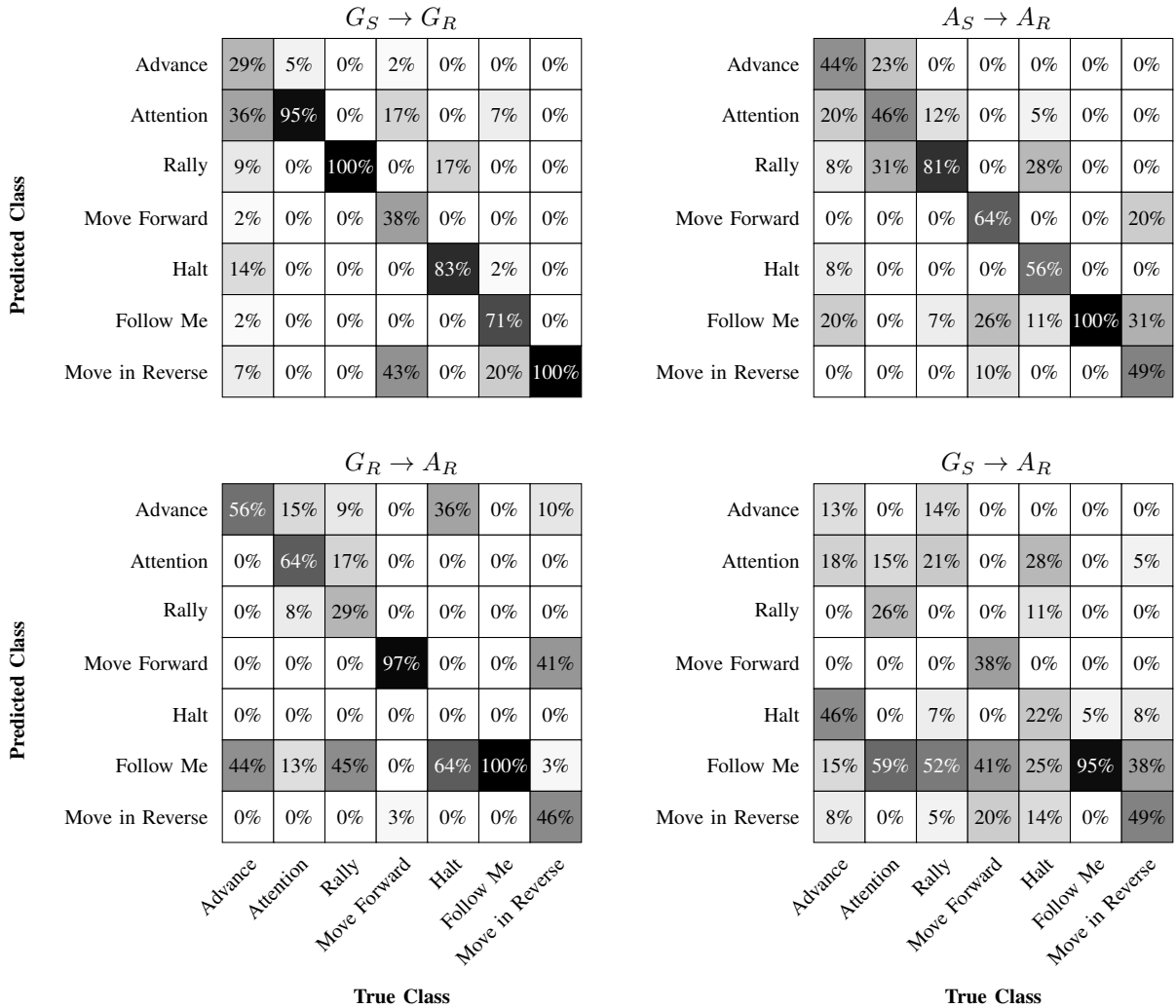
\begin{figure*}
\begin{tikzpicture}[
    scale = \myScale,
    font=\footnotesize
    ]
\tikzset{vertical label/.style={rotate=90,anchor=east}}   
\tikzset{diagonal label/.style={rotate=45,anchor=north east}}

\foreach \y in {1,...,\numClasses} 
{
    \node [anchor=east] at (0.4,-\y) {\pgfmathparse{\classNames[\y-1]}\pgfmathresult}; 
    
    \foreach \x in {1,...,\numClasses}  
    {
    \def\totSamples{0}
    \foreach \ll in {1,...,\numClasses}
    {
        \pgfmathparse{\myConfMat[\ll-1][\x-1]}   
        \xdef\totSamples{\totSamples+\pgfmathresult} 
    }
    \pgfmathparse{\totSamples} \xdef\totSamples{\pgfmathresult}  
    
    \begin{scope}[shift={(\x,-\y)}]
        \def\mVal{\myConfMat[\y-1][\x-1]} 
        \pgfmathtruncatemacro{\r}{\mVal}   %
        \pgfmathtruncatemacro{\p}{round(\r/\totSamples*100)}
        \coordinate (C) at (0,0);
        \ifthenelse{\p<50}{\def\txtcol{black}}{\def\txtcol{white}} 
        \node[
            draw,                 
            text=\txtcol,         
            align=center,         
            fill=black!\p,        
            minimum size=\myScale*10mm,    
            inner sep=0,          
            ] (C) {\p\%};     
        \ifthenelse{\y=\numClasses}{
        {\pgfmathparse{\classNames[\x-1]}\pgfmathresult};}{}
    \end{scope}
    }
}
\coordinate (yaxis) at (-3.5,0.5-\numClasses/2);  
\coordinate (xaxis) at (0.5+\numClasses/2, -\numClasses-1.25); 
\node [vertical label] at (yaxis) {\textbf{Predicted Class}};
\node []               at (xaxis) {\textbf{}};
\node[above, font=\bfseries, xshift=1.5cm] at (current bounding box.north) {$G_S \rightarrow G_R$};
\end{tikzpicture}
\def\myConfMat{{
{436,	231,	0,	0,	0,	0,	0},
{205,	462,	119,	0,	55,	0,	0},
{77,	308,	810,	0,	278,	0,	0},
{0,	0,	0,	641,	0,	0,	205},
{77,	0,	0,	0,	556,	0,	0},
{205,	0,	71,	257,	111,	1000,	308},
{0,	0,	0,	103,	0,	0,	487},
}}
\begin{tikzpicture}[
    scale = \myScale,
    font=\footnotesize
    ]

\tikzset{vertical label/.style={rotate=90,anchor=east}}   
\tikzset{diagonal label/.style={rotate=45,anchor=north east}}

\foreach \y in {1,...,\numClasses} 
{
    \node [anchor=east] at (0.4,-\y) {\pgfmathparse{\classNames[\y-1]}\pgfmathresult}; 
    
    \foreach \x in {1,...,\numClasses}  
    {
    \def\totSamples{0}
    \foreach \ll in {1,...,\numClasses}
    {
        \pgfmathparse{\myConfMat[\ll-1][\x-1]}   
        \xdef\totSamples{\totSamples+\pgfmathresult} 
    }
    \pgfmathparse{\totSamples} \xdef\totSamples{\pgfmathresult}  
    
    \begin{scope}[shift={(\x,-\y)}]
        \def\mVal{\myConfMat[\y-1][\x-1]} 
        \pgfmathtruncatemacro{\r}{\mVal}   %
        \pgfmathtruncatemacro{\p}{round(\r/\totSamples*100)}
        \coordinate (C) at (0,0);
        \ifthenelse{\p<50}{\def\txtcol{black}}{\def\txtcol{white}} 
        \node[
            draw,                 
            text=\txtcol,         
            align=center,         
            fill=black!\p,        
            minimum size=\myScale*10mm,    
            inner sep=0,          
            ] (C) {\p\%};     
        \ifthenelse{\y=\numClasses}{
        {\pgfmathparse{\classNames[\x-1]}\pgfmathresult};}{}
    \end{scope}
    }
}
\coordinate (yaxis) at (-3.5,0.5-\numClasses/2);  
\coordinate (xaxis) at (0.5+\numClasses/2, -\numClasses-1.25); 
\node [vertical label] at (yaxis) {\textbf{}};
\node []               at (xaxis) {\textbf{}};
\node[above, font=\bfseries, xshift=1.5cm] at (current bounding box.north) {$A_S \rightarrow A_R$};
\end{tikzpicture}
\def\myConfMat{{
{564,154,95,0,361,0,103},  
{0,641,166,0,0,0,0},  
{0,77,286,0,0,0,0},  
{0,0,0,974,0,0,410},  
{0,0,0,0,0,0,0},  
{436,128,453,0,639,1000,26},  
{0,0,0,26,0,0,462},  
}}

\begin{tikzpicture}[
    scale = \myScale,
    font=\footnotesize
    ]

\tikzset{vertical label/.style={rotate=90,anchor=east}}   
\tikzset{diagonal label/.style={rotate=45,anchor=north east}}

\foreach \y in {1,...,\numClasses} 
{
    \node [anchor=east] at (0.4,-\y) {\pgfmathparse{\classNames[\y-1]}\pgfmathresult}; 
    
    \foreach \x in {1,...,\numClasses}  
    {
    \def\totSamples{0}
    \foreach \ll in {1,...,\numClasses}
    {
        \pgfmathparse{\myConfMat[\ll-1][\x-1]}   
        \xdef\totSamples{\totSamples+\pgfmathresult} 
    }
    \pgfmathparse{\totSamples} \xdef\totSamples{\pgfmathresult}  
    
    \begin{scope}[shift={(\x,-\y)}]
        \def\mVal{\myConfMat[\y-1][\x-1]} 
        \pgfmathtruncatemacro{\r}{\mVal}   %
        \pgfmathtruncatemacro{\p}{round(\r/\totSamples*100)}
        \coordinate (C) at (0,0);
        \ifthenelse{\p<50}{\def\txtcol{black}}{\def\txtcol{white}} 
        \node[
            draw,                 
            text=\txtcol,         
            align=center,         
            fill=black!\p,        
            minimum size=\myScale*10mm,    
            inner sep=0,          
            ] (C) {\p\%};     
        \ifthenelse{\y=\numClasses}{
        \node [diagonal label] at ($(C)-(0,0.5)$) 
        {\pgfmathparse{\classNames[\x-1]}\pgfmathresult};}{}
    \end{scope}
    }
}
\coordinate (yaxis) at (-3.5,0.5-\numClasses/2);  
\coordinate (xaxis) at (0.5+\numClasses/2, -\numClasses-3.5); 
\node [vertical label] at (yaxis) {\textbf{Predicted Class}};
\node []               at (xaxis) {\textbf{True Class}};
\node[above, font=\bfseries, xshift=1.5cm] at (current bounding box.north) {$G_R \rightarrow A_R$};
\end{tikzpicture}
\def\myConfMat{{
{128,	0,	143,	0,	0,	0,	0},
{180,	154,	214,	0,	278,	0,	51},
{0,	257,	0,	0,	111,	0,	0},
{0,	0,	0,	385,	0,	0,	0},
{462,	0,	71,	0,	222,	51,	77},
{154,	589,	524,	411,	250,	949,	385},
{77,	0,	47,	205,	139,	0,	487},
}}
\begin{tikzpicture}[
    scale = \myScale,
    font=\footnotesize
    ]

\tikzset{vertical label/.style={rotate=90,anchor=east}}   
\tikzset{diagonal label/.style={rotate=45,anchor=north east}}

\foreach \y in {1,...,\numClasses} 
{
    \node [anchor=east] at (0.4,-\y) {\pgfmathparse{\classNames[\y-1]}\pgfmathresult}; 
    
    \foreach \x in {1,...,\numClasses}  
    {
    \def\totSamples{0}
    \foreach \ll in {1,...,\numClasses}
    {
        \pgfmathparse{\myConfMat[\ll-1][\x-1]}   
        \xdef\totSamples{\totSamples+\pgfmathresult} 
    }
    \pgfmathparse{\totSamples} \xdef\totSamples{\pgfmathresult}  
    
    \begin{scope}[shift={(\x,-\y)}]
        \def\mVal{\myConfMat[\y-1][\x-1]} 
        \pgfmathtruncatemacro{\r}{\mVal}   %
        \pgfmathtruncatemacro{\p}{round(\r/\totSamples*100)}
        \coordinate (C) at (0,0);
        \ifthenelse{\p<50}{\def\txtcol{black}}{\def\txtcol{white}} 
        \node[
            draw,                 
            text=\txtcol,         
            align=center,         
            fill=black!\p,        
            minimum size=\myScale*10mm,    
            inner sep=0,          
            ] (C) {\p\%};     
        \ifthenelse{\y=\numClasses}{
        \node [diagonal label] at ($(C)-(0,0.5)$) 
        {\pgfmathparse{\classNames[\x-1]}\pgfmathresult};}{}
    \end{scope}
    }
}
\coordinate (yaxis) at (-3.5,0.5-\numClasses/2);  
\coordinate (xaxis) at (0.5+\numClasses/2, -\numClasses-3.5); 
\node [vertical label] at (yaxis) {\textbf{}};
\node []               at (xaxis) {\textbf{True Class}};
\node[above,font=\bfseries,xshift=1.5cm] at (current bounding box.north) {$G_S \rightarrow A_R$};
\end{tikzpicture}

\caption{Confusion matrices for action recognition in the four UDA settings using CO\textsuperscript{2}A with X3D backbone (averaged across three runs). Top Left: $G_S \rightarrow G_R$, Top Right: $A_S \rightarrow A_R$, Bottom Left: $G_R \rightarrow A_R$, Bottom Right: $G_S \rightarrow A_R$.}
\label{fig:confusion}
\end{figure*}

\section{Conclusion}
\label{sec:conclusion}

Domain adaptation, which seeks to address the effects of distributional shift that occur when DNNs are applied in domains with different characteristics than those used for training, is central to the success of many robotics applications. Given the practical challenges of acquiring high-quality, real-world labeled data and the proliferation of synthetic data, addressing synthetic-to-real domain shift becomes essential ~\cite{deMeloEtAl22}. Here, we present a novel dataset, with comparable real and synthetic data, to support the development and comparison of DA algorithms for action recognition. To further understand the generalization of algorithms, we provide data acquired from different viewing perspectives -- ground and air. Our baselines reinforce the promise of synthetic data for deep learning while still replicating the typical synthetic-to-real gap. The dataset lends itself to several potentially interesting follow up studies. For instance, comparing animated vs. mocap synthetic data could provide insight on the importance of motion realism and ablation studies could shed light on the relative contribution of each synthesis parameter. Follow-up experimentation could also try to tease apart which aspects of the synthetic data are contributing to performance the most. Our baseline experiments also confirm the existence of a ground-to-air domain gap. Here too, synthetic data was helpful, as training from data generated for the appropriate viewing perspective led to a boost in performance in some cases. 

The richness of the dataset introduces considerable opportunities for further study. While we focus here on pure RGB-based approaches, researchers have shown the value in considering other modalities for action recognition, such as motion flow and skeleton estimation ~\cite{carreira2017quo, choutas2018potion, weinzaepfel2021mimetics, yan2019pa3d}. Scene segmentation information, which is also available for the synthetic data, was shown to be useful in closing the synthetic-to-real gap ~\cite{hoffman2018cycada}. Prior work has further suggested that mixing synthetic and real data, using various fine tuning strategies, can lead to better performance than using either of the data types alone ~\cite{deMeloEtAl20, deSouzaEtAl17, GaidonEtAl16, ShafaeiLittle16}. While our baseline experiments focused on the most typical cases, the synthetic-to-real and ground-to-air domain shifts, it is possible to consider other transfer scenarios. These include the reverse problems of shifting from the air to the ground perspective or shifting from  real to synthetic (in order to leverage perception pipelines that have been optimized using synthetic data). Finally, any algorithm will ultimately have to be deployed on robotic platforms with limited compute and power. It is, thus, important to complement the baselines reported here with analyses of the real-time performance of algorithms when deployed in a variety of robotic platforms.

\section{Acknowledgments}
\label{sec:acknowledgments}

This work was supported by Army Research Laboratory (ARL) Cooperative Agreements W911NF-21-2-0211 and W911NF-21-2-0076. We would like to thank our collaborators at the University of Maryland for their invaluable help in the data collection efforts: Darren William Robey, Grant Williams, and Josh Gaus.

\clearpage
\bibliographystyle{IEEEtran}
\bibliography{references}

\begin{thebibliography}{10}
\providecommand{\url}[1]{#1}
\csname url@samestyle\endcsname
\providecommand{\newblock}{\relax}
\providecommand{\bibinfo}[2]{#2}
\providecommand{\BIBentrySTDinterwordspacing}{\spaceskip=0pt\relax}
\providecommand{\BIBentryALTinterwordstretchfactor}{4}
\providecommand{\BIBentryALTinterwordspacing}{\spaceskip=\fontdimen2\font plus
\BIBentryALTinterwordstretchfactor\fontdimen3\font minus
  \fontdimen4\font\relax}
\providecommand{\BIBforeignlanguage}[2]{{%
\expandafter\ifx\csname l@#1\endcsname\relax
\typeout{** WARNING: IEEEtran.bst: No hyphenation pattern has been}%
\typeout{** loaded for the language `#1'. Using the pattern for}%
\typeout{** the default language instead.}%
\else
\language=\csname l@#1\endcsname
\fi
#2}}
\providecommand{\BIBdecl}{\relax}
\BIBdecl

\bibitem{visda2017}
X.~Peng, B.~Usman, N.~Kaushik, J.~Hoffman, D.~Wang, and K.~Saenko, ``Visda: The
  visual domain adaptation challenge,'' 2017.

\bibitem{simonyan2014two}
K.~Simonyan and A.~Zisserman, ``Two-stream convolutional networks for action
  recognition in videos,'' \emph{Advances in neural information processing
  systems}, vol.~27, 2014.

\bibitem{karpathy2014large}
A.~Karpathy, G.~Toderici, S.~Shetty, T.~Leung, R.~Sukthankar, and L.~Fei-Fei,
  ``Large-scale video classification with convolutional neural networks,'' in
  \emph{Proceedings of the IEEE conference on Computer Vision and Pattern
  Recognition}, 2014, pp. 1725--1732.

\bibitem{negin2016human}
F.~Negin and F.~Bremond, ``Human action recognition in videos: A survey,''
  \emph{INRIA Technical Report}, 2016.

\bibitem{yue2015beyond}
J.~Yue-Hei~Ng, M.~Hausknecht, S.~Vijayanarasimhan, O.~Vinyals, R.~Monga, and
  G.~Toderici, ``Beyond short snippets: Deep networks for video
  classification,'' in \emph{Proceedings of the IEEE conference on computer
  vision and pattern recognition}, 2015, pp. 4694--4702.

\bibitem{carreira2017quo}
J.~Carreira and A.~Zisserman, ``Quo vadis, action recognition? a new model and
  the kinetics dataset,'' in \emph{proceedings of the IEEE Conference on
  Computer Vision and Pattern Recognition}, 2017, pp. 6299--6308.

\bibitem{hou2017tube}
R.~Hou, C.~Chen, and M.~Shah, ``Tube convolutional neural network (t-cnn) for
  action detection in videos,'' in \emph{Proceedings of the IEEE international
  conference on computer vision}, 2017, pp. 5822--5831.

\bibitem{saha2017amtnet}
S.~Saha, G.~Singh, and F.~Cuzzolin, ``Amtnet: Action-micro-tube regression by
  end-to-end trainable deep architecture,'' in \emph{Proceedings of the IEEE
  International Conference on Computer Vision}, 2017, pp. 4414--4423.

\bibitem{choutas2018potion}
V.~Choutas, P.~Weinzaepfel, J.~Revaud, and C.~Schmid, ``Potion: Pose motion
  representation for action recognition,'' in \emph{Proceedings of the IEEE
  conference on computer vision and pattern recognition}, 2018, pp. 7024--7033.

\bibitem{weinzaepfel2021mimetics}
P.~Weinzaepfel and G.~Rogez, ``Mimetics: Towards understanding human actions
  out of context,'' \emph{International Journal of Computer Vision}, vol. 129,
  no.~5, pp. 1675--1690, 2021.

\bibitem{yan2019pa3d}
A.~Yan, Y.~Wang, Z.~Li, and Y.~Qiao, ``Pa3d: Pose-action 3d machine for video
  recognition,'' in \emph{Proceedings of the IEEE/CVF Conference on Computer
  Vision and Pattern Recognition}, 2019, pp. 7922--7931.

\bibitem{saenko2010adapting}
K.~Saenko, B.~Kulis, M.~Fritz, and T.~Darrell, ``Adapting visual category
  models to new domains,'' in \emph{European conference on computer
  vision}.\hskip 1em plus 0.5em minus 0.4em\relax Springer, 2010, pp. 213--226.

\bibitem{peng2019moment}
X.~Peng, Q.~Bai, X.~Xia, Z.~Huang, K.~Saenko, and B.~Wang, ``Moment matching
  for multi-source domain adaptation,'' in \emph{Proceedings of the IEEE/CVF
  international conference on computer vision}, 2019, pp. 1406--1415.

\bibitem{sankaranarayanan2018learning}
S.~Sankaranarayanan, Y.~Balaji, A.~Jain, S.~N. Lim, and R.~Chellappa,
  ``Learning from synthetic data: Addressing domain shift for semantic
  segmentation,'' in \emph{Proceedings of the IEEE conference on computer
  vision and pattern recognition}, 2018, pp. 3752--3761.

\bibitem{hoffman2018cycada}
J.~Hoffman, E.~Tzeng, T.~Park, J.-Y. Zhu, P.~Isola, K.~Saenko, A.~Efros, and
  T.~Darrell, ``Cycada: Cycle-consistent adversarial domain adaptation,'' in
  \emph{International conference on machine learning}.\hskip 1em plus 0.5em
  minus 0.4em\relax Pmlr, 2018, pp. 1989--1998.

\bibitem{deMeloEtAl22}
C.~M. de~Melo, A.~Torralba, L.~Guibas, J.~DiCarlo, R.~Chellappa, and
  J.~Hodgins, ``Next-generation deep learning based on simulators and synthetic
  data,'' \emph{Trends in Cognitive Sciences}, vol.~26, pp. 174--187, 2021.

\bibitem{Nikolenko19}
S.~Nikolenko, ``Synthetic data for deep learning,'' 2019.

\bibitem{9649285}
L.~Zherdeva, E.~Minaev, D.~Zherdev, and V.~Fursov, ``Synthetic dataset for
  navigation tasks of autonomous systems and ground robots,'' in \emph{2021
  International Conference on Information Technology and Nanotechnology
  (ITNT)}, 2021, pp. 1--4.

\bibitem{10.1016/j.robot.2019.103336}
\BIBentryALTinterwordspacing
S.~Wang, J.~Yue, Y.~Dong, S.~He, H.~Wang, and S.~Ning, ``A synthetic dataset
  for visual slam evaluation,'' \emph{Robot. Auton. Syst.}, vol. 124, no.~C,
  feb 2020. [Online]. Available:
  \url{https://doi.org/10.1016/j.robot.2019.103336}
\BIBentrySTDinterwordspacing

\bibitem{7487210}
Z.~Zhang, H.~Rebecq, C.~Forster, and D.~Scaramuzza, ``Benefit of large
  field-of-view cameras for visual odometry,'' in \emph{2016 IEEE International
  Conference on Robotics and Automation (ICRA)}, 2016, pp. 801--808.

\bibitem{DBLP:journals/corr/abs-2112-12252}
\BIBentryALTinterwordspacing
B.~Kiefer, D.~Ott, and A.~Zell, ``Leveraging synthetic data in object detection
  on unmanned aerial vehicles,'' \emph{CoRR}, vol. abs/2112.12252, 2021.
  [Online]. Available: \url{https://arxiv.org/abs/2112.12252}
\BIBentrySTDinterwordspacing

\bibitem{DBLP:journals/corr/abs-2007-11118}
\BIBentryALTinterwordspacing
O.~Matthews, K.~Ryu, and T.~Srivastava, ``Creating a large-scale synthetic
  dataset for human activity recognition,'' \emph{CoRR}, vol. abs/2007.11118,
  2020. [Online]. Available: \url{https://arxiv.org/abs/2007.11118}
\BIBentrySTDinterwordspacing

\bibitem{9206624}
F.~Alharbi, L.~Ouarbya, and J.~A. Ward, ``Synthetic sensor data for human
  activity recognition,'' in \emph{2020 International Joint Conference on
  Neural Networks (IJCNN)}, 2020, pp. 1--9.

\bibitem{9340728}
C.~M. de~Melo, B.~Rothrock, P.~Gurram, O.~Ulutan, and B.~Manjunath,
  ``Vision-based gesture recognition in human-robot teams using synthetic
  data,'' in \emph{2020 IEEE/RSJ International Conference on Intelligent Robots
  and Systems (IROS)}, 2020, pp. 10\,278--10\,284.

\bibitem{nguyen2022state}
K.~Nguyen, C.~Fookes, S.~Sridharan, Y.~Tian, X.~Liu, F.~Liu, and A.~Ross, ``The
  state of aerial surveillance: A survey,'' \emph{arXiv preprint
  arXiv:2201.03080}, 2022.

\bibitem{li2021uav}
T.~Li, J.~Liu, W.~Zhang, Y.~Ni, W.~Wang, and Z.~Li, ``Uav-human: A large
  benchmark for human behavior understanding with unmanned aerial vehicles,''
  in \emph{Proceedings of the IEEE/CVF conference on computer vision and
  pattern recognition}, 2021, pp. 16\,266--16\,275.

\bibitem{perera2018uav}
A.~G. Perera, Y.~Wei~Law, and J.~Chahl, ``Uav-gesture: A dataset for uav
  control and gesture recognition,'' in \emph{Proceedings of the European
  Conference on Computer Vision (ECCV) Workshops}, 2018, pp. 0--0.

\bibitem{barekatain2017okutama}
M.~Barekatain, M.~Mart{\'\i}, H.-F. Shih, S.~Murray, K.~Nakayama, Y.~Matsuo,
  and H.~Prendinger, ``Okutama-action: An aerial view video dataset for
  concurrent human action detection,'' in \emph{Proceedings of the IEEE
  conference on computer vision and pattern recognition workshops}, 2017, pp.
  28--35.

\bibitem{choi2020unsupervised}
J.~Choi, G.~Sharma, M.~Chandraker, and J.-B. Huang, ``Unsupervised and
  semi-supervised domain adaptation for action recognition from drones,'' in
  \emph{Proceedings of the IEEE/CVF Winter Conference on Applications of
  Computer Vision}, 2020, pp. 1717--1726.

\bibitem{Dosovitskiy17}
A.~Dosovitskiy, G.~Ros, F.~Codevilla, A.~Lopez, and V.~Koltun, ``{CARLA}: {An}
  open urban driving simulator,'' in \emph{Proceedings of the 1st Annual
  Conference on Robot Learning}, 2017, pp. 1--16.

\bibitem{Richter2016PlayingFD}
S.~R. Richter, V.~Vineet, S.~Roth, and V.~Koltun, ``Playing for data: Ground
  truth from computer games,'' \emph{ArXiv}, vol. abs/1608.02192, 2016.

\bibitem{2021_28dd2c79}
\BIBentryALTinterwordspacing
V.~Makoviychuk, L.~Wawrzyniak, Y.~Guo, M.~Lu, K.~Storey, M.~Macklin,
  D.~Hoeller, N.~Rudin, A.~Allshire, A.~Handa, and G.~State, ``Isaac gym: High
  performance gpu based physics simulation for robot learning,'' in
  \emph{Proceedings of the Neural Information Processing Systems Track on
  Datasets and Benchmarks}, J.~Vanschoren and S.~Yeung, Eds., vol.~1, 2021.
  [Online]. Available:
  \url{https://datasets-benchmarks-proceedings.neurips.cc/paper/2021/file/28dd2c7955ce926456240b2ff0100bde-Paper-round2.pdf}
\BIBentrySTDinterwordspacing

\bibitem{szot2021habitat}
A.~Szot, A.~Clegg, E.~Undersander, E.~Wijmans, Y.~Zhao, J.~Turner, N.~Maestre,
  M.~Mukadam, D.~Chaplot, O.~Maksymets, A.~Gokaslan, V.~Vondrus, S.~Dharur,
  F.~Meier, W.~Galuba, A.~Chang, Z.~Kira, V.~Koltun, J.~Malik, M.~Savva, and
  D.~Batra, ``Habitat 2.0: Training home assistants to rearrange their
  habitat,'' in \emph{Advances in Neural Information Processing Systems
  (NeurIPS)}, 2021.

\bibitem{habitat19iccv}
M.~Savva, A.~Kadian, O.~Maksymets, Y.~Zhao, E.~Wijmans, B.~Jain, J.~Straub,
  J.~Liu, V.~Koltun, J.~Malik, D.~Parikh, and D.~Batra, ``Habitat: {A}
  {P}latform for {E}mbodied {AI} {R}esearch,'' in \emph{Proceedings of the
  IEEE/CVF International Conference on Computer Vision (ICCV)}, 2019.

\bibitem{RoomR}
L.~Weihs, M.~Deitke, A.~Kembhavi, and R.~Mottaghi, ``Visual room
  rearrangement,'' in \emph{IEEE/CVF Conference on Computer Vision and Pattern
  Recognition (CVPR)}, June 2021.

\bibitem{li2021igibson}
\BIBentryALTinterwordspacing
C.~Li, F.~Xia, R.~Mart{\'\i}n-Mart{\'\i}n, M.~Lingelbach, S.~Srivastava,
  B.~Shen, K.~E. Vainio, C.~Gokmen, G.~Dharan, T.~Jain, A.~Kurenkov, K.~Liu,
  H.~Gweon, J.~Wu, L.~Fei-Fei, and S.~Savarese, ``igibson 2.0: Object-centric
  simulation for robot learning of everyday household tasks,'' in \emph{5th
  Annual Conference on Robot Learning}, 2021. [Online]. Available:
  \url{https://openreview.net/forum?id=2uGN5jNJROR}
\BIBentrySTDinterwordspacing

\bibitem{10.1007/978-3-031-04870-8_16}
K.~Dimitropoulos, I.~Hatzilygeroudis, and K.~Chatzilygeroudis, ``A brief survey
  of sim2real methods for robot learning,'' in \emph{Advances in Service and
  Industrial Robotics}, A.~M{\"u}ller and M.~Brandst{\"o}tter, Eds.\hskip 1em
  plus 0.5em minus 0.4em\relax Cham: Springer International Publishing, 2022,
  pp. 133--140.

\bibitem{DBLP:journals/corr/abs-2012-03806}
\BIBentryALTinterwordspacing
S.~H{\"{o}}fer, K.~E. Bekris, A.~Handa, J.~C.~G. Higuera, F.~Golemo,
  M.~Mozifian, C.~G. Atkeson, D.~Fox, K.~Goldberg, J.~Leonard, C.~K. Liu,
  J.~Peters, S.~Song, P.~Welinder, and M.~White, ``Perspectives on sim2real
  transfer for robotics: {A} summary of the {R:} {SS} 2020 workshop,''
  \emph{CoRR}, vol. abs/2012.03806, 2020. [Online]. Available:
  \url{https://arxiv.org/abs/2012.03806}
\BIBentrySTDinterwordspacing

\bibitem{10.5555/3045118.3045130}
M.~Long, Y.~Cao, J.~Wang, and M.~I. Jordan, ``Learning transferable features
  with deep adaptation networks,'' in \emph{Proceedings of the 32nd
  International Conference on International Conference on Machine Learning -
  Volume 37}, ser. ICML'15.\hskip 1em plus 0.5em minus 0.4em\relax JMLR.org,
  2015, p. 97–105.

\bibitem{10.1145/3357384.3357918}
\BIBentryALTinterwordspacing
M.~Cao, X.~Zhou, Y.~Xu, Y.~Pang, and B.~Yao, ``Adversarial domain adaptation
  with semantic consistency for cross-domain image classification,'' in
  \emph{Proceedings of the 28th ACM International Conference on Information and
  Knowledge Management}, ser. CIKM '19.\hskip 1em plus 0.5em minus 0.4em\relax
  New York, NY, USA: Association for Computing Machinery, 2019, p. 259–268.
  [Online]. Available: \url{https://doi.org/10.1145/3357384.3357918}
\BIBentrySTDinterwordspacing

\bibitem{8578985}
S.~Sankaranarayanan, Y.~Balaji, C.~D. Castillo, and R.~Chellappa, ``Generate to
  adapt: Aligning domains using generative adversarial networks,'' in
  \emph{2018 IEEE/CVF Conference on Computer Vision and Pattern Recognition},
  2018, pp. 8503--8512.

\bibitem{inproceedingsgrasp}
S.~James, P.~Wohlhart, M.~Kalakrishnan, D.~Kalashnikov, A.~Irpan, J.~Ibarz,
  S.~Levine, R.~Hadsell, and K.~Bousmalis, ``Sim-to-real via sim-to-sim:
  Data-efficient robotic grasping via randomized-to-canonical adaptation
  networks,'' 06 2019, pp. 12\,619--12\,629.

\bibitem{Bousmalis2018UsingSA}
K.~Bousmalis, A.~Irpan, P.~Wohlhart, Y.~Bai, M.~Kelcey, M.~Kalakrishnan,
  L.~Downs, J.~Ibarz, P.~Pastor, K.~Konolige, S.~Levine, and V.~Vanhoucke,
  ``Using simulation and domain adaptation to improve efficiency of deep
  robotic grasping,'' \emph{2018 IEEE International Conference on Robotics and
  Automation (ICRA)}, pp. 4243--4250, 2018.

\bibitem{9561157}
D.~Ho, K.~Rao, Z.~Xu, E.~Jang, M.~Khansari, and Y.~Bai, ``Retinagan: An
  object-aware approach to sim-to-real transfer,'' in \emph{2021 IEEE
  International Conference on Robotics and Automation (ICRA)}, 2021, pp.
  10\,920--10\,926.

\bibitem{8793789}
Y.~Chebotar, A.~Handa, V.~Makoviychuk, M.~Macklin, J.~Issac, N.~Ratliff, and
  D.~Fox, ``Closing the sim-to-real loop: Adapting simulation randomization
  with real world experience,'' in \emph{2019 International Conference on
  Robotics and Automation (ICRA)}, 2019, pp. 8973--8979.

\bibitem{8793561}
J.~v. Baar, A.~Sullivan, R.~Cordorel, D.~Jha, D.~Romeres, and D.~Nikovski,
  ``Sim-to-real transfer learning using robustified controllers in robotic
  tasks involving complex dynamics,'' in \emph{2019 International Conference on
  Robotics and Automation (ICRA)}, 2019, pp. 6001--6007.

\bibitem{DBLP:journals/ral/ZhangTYX0BB19}
J.~Zhang, L.~Tai, P.~Yun, Y.~Xiong, M.~Liu, J.~Boedecker, and W.~Burgard,
  ``Vr-goggles for robots: Real-to-sim domain adaptation for visual control,''
  \emph{{IEEE} Robotics Autom. Lett.}, vol.~4, no.~2, pp. 1148--1155, 2019.

\bibitem{sim2real_outdoor}
K.~Weerakoon, A.~Sathyamoorthy, and D.~Manocha, ``Sim-to-real strategy for
  spatially aware robot navigation in uneven outdoor environments,'' 05 2022.

\bibitem{DANN}
Y.~Ganin, E.~Ustinova, H.~Ajakan, P.~Germain, H.~Larochelle, F.~Laviolette,
  M.~Marchand, and V.~Lempitsky, ``Domain-adversarial training of neural
  networks,'' \emph{The journal of machine learning research}, vol.~17, no.~1,
  pp. 2096--2030, 2016.

\bibitem{CO2A}
V.~G.~T. da~Costa, G.~Zara, P.~Rota, T.~Oliveira-Santos, N.~Sebe, V.~Murino,
  and E.~Ricci, ``Dual-head contrastive domain adaptation for video action
  recognition,'' in \emph{Proceedings of the IEEE/CVF Winter Conference on
  Applications of Computer Vision}, 2022, pp. 1181--1190.

\bibitem{deMeloEtAl20}
C.~de~Melo, B.~Rothrock, O.~U. P.~Gurram, and B.~Manjunath, ``Vision-based
  gesture recognition in human-robot teams using synthetic data,'' in
  \emph{Proc. of the IROS'20}, 2020.

\bibitem{USArmy87}
\BIBentryALTinterwordspacing
``Visual signals: Field manual 21-60,'' 1987. [Online]. Available:
  \url{https://www.radford.edu/content/dam/colleges/chbs/rotc/Forms/fm/Visual\%20Signals\%20FM\%2021-60.pdf}
\BIBentrySTDinterwordspacing

\bibitem{wu2019detectron2}
Y.~Wu, A.~Kirillov, F.~Massa, W.-Y. Lo, and R.~Girshick, ``Detectron2,''
  \url{https://github.com/facebookresearch/detectron2}, 2019.

\bibitem{2020mmaction2}
M.~Contributors, ``Openmmlab's next generation video understanding toolbox and
  benchmark,'' \url{https://github.com/open-mmlab/mmaction2}, 2020.

\bibitem{feichtenhofer2020x3d}
C.~Feichtenhofer, ``X3d: Expanding architectures for efficient video
  recognition,'' in \emph{Proceedings of the IEEE/CVF Conference on Computer
  Vision and Pattern Recognition}, 2020, pp. 203--213.

\bibitem{NEURIPS2020_d85b63ef}
\BIBentryALTinterwordspacing
E.~D. Cubuk, B.~Zoph, J.~Shlens, and Q.~Le, ``Randaugment: Practical automated
  data augmentation with a reduced search space,'' in \emph{Advances in Neural
  Information Processing Systems}, H.~Larochelle, M.~Ranzato, R.~Hadsell,
  M.~Balcan, and H.~Lin, Eds., vol.~33.\hskip 1em plus 0.5em minus 0.4em\relax
  Curran Associates, Inc., 2020, pp. 18\,613--18\,624. [Online]. Available:
  \url{https://proceedings.neurips.cc/paper/2020/file/d85b63ef0ccb114d0a3bb7b7d808028f-Paper.pdf}
\BIBentrySTDinterwordspacing

\bibitem{deSouzaEtAl17}
C.~de~Souza, A.~Gaidon, Y.~Cabon, and A.~López, ``Procedural generation of
  videos to train deep action recognition networks,'' in \emph{Proc. of
  CVPR'17}, 2017.

\bibitem{GaidonEtAl16}
A.~Gaidon, Q.~Wang, Y.~Cabon, and E.~Vig, ``Virtual worlds as proxy for
  multi-object tracking analysis,'' in \emph{Proc. of CVPR'16}, 2016.

\bibitem{ShafaeiLittle16}
A.~Shafaei and J.~Little, ``Real-time human motion capture with multiple depth
  cameras,'' in \emph{Proc. of CRV'16}, 2016.

\end{thebibliography}

\end{document}